\pdfoutput=1
\documentclass{article}

\PassOptionsToPackage{numbers,compress}{natbib}
\usepackage[final,main]{neurips_2026}
\usepackage[utf8]{inputenc}
\usepackage[T1]{fontenc}
\usepackage{hyperref}
\hypersetup{hidelinks}
\usepackage{url}
\usepackage{booktabs}
\usepackage{amsfonts}
\usepackage{amsmath}
\usepackage{amssymb}
\usepackage{nicefrac}
\usepackage{microtype}
\usepackage{xcolor}
\usepackage{colortbl}
\usepackage[most]{tcolorbox}
\usepackage{graphicx}
\usepackage{wrapfig}
\usepackage{capt-of}
\usepackage{multirow}
\usepackage{makecell}
\usepackage{array}
\usepackage{tabularx}
\usepackage{enumitem}
\usepackage{tikz}
\usepackage{pgfplots}
\usepackage{pifont}
\usepackage{titletoc}
\usepackage[capitalize,nameinlink]{cleveref}

\newcommand{\cmark}{\ding{51}}
\newcommand{\xmark}{\ding{55}}

\tcbset{
  promptbox/.style={
    width=\linewidth,
    top=8pt,
    bottom=6pt,
    left=8pt,
    right=8pt,
    colback=white,
    colframe=hmntakeawaygreen,
    colbacktitle=hmntakeawaygreen,
    coltitle=white,
    enhanced,
    boxrule=0.6pt,
    arc=3pt,
    fonttitle=\small\bfseries,
    attach boxed title to top left={yshift=-0.1in,xshift=0.15in},
    boxed title style={boxrule=0pt,colframe=white},
  },
}
\newtcolorbox{promptbox}[2][]{promptbox,title=#2,#1}

\usetikzlibrary{arrows.meta,fit,positioning}
\pgfplotsset{compat=1.18}
\setlist[itemize]{leftmargin=*, topsep=2pt, itemsep=1pt}

\definecolor{hmntakeawaygreen}{rgb}{0.18, 0.55, 0.34}
\definecolor{box_green}{RGB}{0, 180, 139}
\definecolor{box_red}{RGB}{239, 99, 75}
\definecolor{fig_green}{HTML}{82B366}
\definecolor{fig_red}{HTML}{B85450}
\definecolor{fig_grey}{HTML}{6E6E6E}
\newtcolorbox{hmntakeaway}[1]{
  enhanced,
  colback=hmntakeawaygreen!8,
  colframe=hmntakeawaygreen,
  boxrule=0.7pt,
  arc=5pt,
  left=3pt,
  right=3pt,
  top=3pt,
  bottom=3pt,
  fontupper=\footnotesize\raggedright,
  title={#1},
  coltitle=white,
  fonttitle=\bfseries\footnotesize,
  colbacktitle=hmntakeawaygreen!90!black
}

\newcommand{\method}{SMILE}
\newcommand{\circles}{CIRCLES}
\newcommand{\bestscore}[1]{\cellcolor{box_green!12}\textcolor{box_green}{#1}}
\newcommand{\secondscore}[1]{\cellcolor{box_red!12}\textcolor{box_red}{#1}}
\newcommand{\methodtable}{\textbf{\textcolor{box_green}{\method~(ours)}}}

\title{The Value of Being Wrong:\\Self-Mined Visual In-Context Learning from Errors}

\author{%
  Shu Zhao \qquad Tan Yu \\
  NVIDIA \\
  \texttt{shuzhao.cs@gmail.com}
}

\begin{document}

\maketitle

\begin{abstract}
Visual in-context learning (ICL) enables vision-language models (VLMs) to adapt at test time via demonstration examples, but its effectiveness depends critically on the examples selected. Recent counterfactual selection methods improve ICL by constructing demonstrations that expose how individual attribute changes affect the answer. However, they require multi-stage pipelines involving attribute extraction, caption engineering, and composed image matching. We propose a simpler alternative grounded in a key observation: \emph{a model's own prediction errors are natural counterfactual demonstrations that directly encode its failure modes.} Our method, \method{} (\textbf{S}elf-\textbf{M}ined Visual \textbf{I}n-Context \textbf{L}earning from \textbf{E}rrors), builds a model-specific pool of hard negatives offline by recording the target VLM's own mistakes. At test time, the prediction serves as a direction signal that surfaces, in the semantic embedding space, the past mistakes most likely to recur for the current query. This prediction-conditioned identification provides a unified mechanism for both closed-set classification and open-ended visual question answering without task-specific design. Across four benchmarks and four VLMs, \method{} delivers an average task-score gain of $+3.09$ points over the SOTA baseline at $5\times$ lower latency. Project page: \url{https://tree-shu-zhao.github.io/SMILE/}.
\end{abstract}
\section{Introduction}
\label{sec:intro}

In-context learning (ICL) has become the dominant paradigm for adapting vision-language models (VLMs) to new tasks at test time without parameter updates~\citep{alayrac2022flamingo,li2023blip,liu2023visual,tsimpoukelli2021multimodal,awadalla2023openflamingo}, unlike fine-tuning~\citep{zhao2026mora,zhao2025kalahash}, reinforcement learning~\citep{zhao2025parallelsearch}, or training task-specific models~\citep{zhao2020asymmetric,zhao2021rescuing,hao2025mapfusion,hao2025safemap}. However, the quality of the demonstration set matters as much as the model itself: poorly selected examples can reinforce spurious correlations~\citep{harutyunyan2024incontext}, while carefully curated ones can expose the fine-grained distinctions that determine the correct answer~\citep{li2024configure,luo2024textual,kim2024finer}. This sensitivity has spurred active research into demonstration-selection strategies, from nearest-neighbor image matching~\citep{yang2022empirical} to multimodal heuristics~\citep{liu2022makes,lu2022fantastically,balazevic2023towards,chen2023mmices,huang2024multimodal,yi2025drum}.

Recently, \circles~\citep{xiong2026circles} shows that counterfactual demonstration selection, assembling demonstrations that isolate how individual attribute changes affect the answer, helps VLMs reason about fine-grained visual distinctions. However, constructing each counterfactual requires the VLM to extract task-relevant attributes, generate faithful counterfactual captions, and execute composed image matching against the labeled corpus, a multi-stage pipeline with three potential points of failure.

\begin{figure}[!t]
    \centering
    \includegraphics[width=\textwidth]{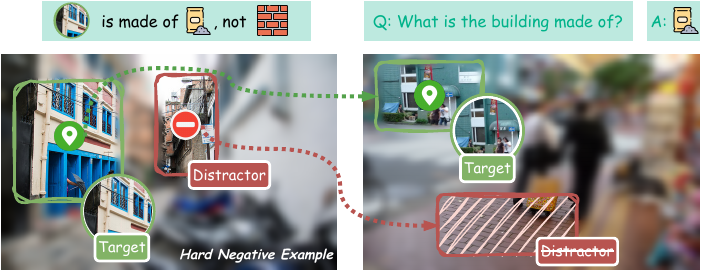}
    \caption{\textbf{A VLM's own past mistakes can steer attention away from visual distractors and onto the target region.} Direct answering and positive-only retrieval focus on visually similar but misleading evidence, while attribute-based counterfactual retrieval fails to isolate the relevant material distinction. \method{} instead retrieves a hard negative with the same error pattern from the mined pool, making the distractor evidence explicit and steering the VLM toward the correct visual region.}
    \label{fig:teaser}
    \vspace{-5mm}
\end{figure}

This paper proposes a simpler path. Our key observation is that \emph{a model's own prediction errors already function as natural counterfactuals}. As illustrated in \Cref{fig:teaser}, a hard negative retrieved from the model's past errors can expose the same visual failure mode as the query, helping the VLM reject a plausible but wrong answer that positive demonstrations or attribute-based retrieval may not disambiguate. When a VLM predicts ``\texttt{brick}'' for a building image whose ground-truth answer is ``\texttt{concrete},'' that example demonstrates that the visual evidence the model considers sufficient for one answer actually corresponds to a different one: precisely what a counterfactual demonstration should convey. Crucially, these errors are not arbitrary: we show empirically (\Cref{sec:analysis}) that classification mistakes recur as exact confusion pairs, while VQA mistakes remain semantically close to the correct answer even when exact strings differ. This structure turns the model's own zero-shot prediction into a useful cue for selecting past errors that resemble its current likely mistake.

Building on this insight, we propose \method{} (\textbf{S}elf-\textbf{M}ined Visual \textbf{I}n-Context \textbf{L}earning from \textbf{E}rrors). An offline mining pass collects incorrectly predicted labeled examples as a reusable hard-negative pool. At test time, positive demonstrations are identified by image-embedding similarity, then the model's zero-shot prediction guides identification of hard negatives whose stored mistaken predictions are semantically aligned in text-embedding space. Across four benchmarks and four VLMs, \method{} delivers an average task-score gain of $+3.09$ points over the SOTA baseline at $5\times$ lower latency.

Our contributions are:
\begin{itemize}
    \item We reformulate counterfactual demonstration selection for visual ICL as model-specific hard-negative mining, a simple yet effective training-free framework.
    \item We provide empirical analysis showing that model errors are a reusable selection signal: exact confusions recur in closed-set classification, semantic error directions persist in open-ended VQA, and a pool decomposition explains why mined hard examples remain useful rather than noisy.
    \item We introduce prediction-conditioned selection via text-embedding similarity, a unified mechanism for both closed-set classification and open-ended VQA, and demonstrate consistent improvements across four benchmarks and four VLMs from two model families.
\end{itemize}

\section{Related Work}
\label{sec:related}

\textbf{Visual in-context learning.}
Vision-language models~\citep{alayrac2022flamingo,li2023blip,liu2023visual} can adapt to new tasks at test time by conditioning on a handful of image--text demonstrations~\citep{tsimpoukelli2021multimodal,alayrac2022flamingo}, but the resulting performance is highly sensitive to the demonstration set~\citep{liu2022makes,lu2022fantastically,rubin2022learning,balazevic2023towards,harutyunyan2024incontext}. RICES picks nearest visual neighbors using CLIP features~\citep{yang2022empirical,awadalla2023openflamingo,radford2021learning}, and subsequent visual ICL work has refined this in several directions: \citet{li2024configure} configure in-context sequences for visual QA, \citet{luo2024textual} introduce the MUIER family that incorporates textual similarity, MMICES uses a two-stage multimodal selector~\citep{chen2023mmices}, \citet{chen2025provoking} explore exploration--exploitation strategies, \citet{huang2024multimodal,zhou2024visual,yi2025drum} learn or curate selection policies tailored to multimodal ICL, \citet{chen2025enhancing} apply coreset optimization, and \citet{chen2024understanding} analyze the mechanisms behind visual ICL effectiveness. These approaches share a common thread: they optimize example selection to match the query. \circles~\citep{xiong2026circles} departs from this paradigm with \emph{counterfactual} demonstrations constructed via attribute extraction and composed image matching. Our method takes a fundamentally different approach: rather than \emph{synthesizing} counterfactuals via attribute interventions, we leverage the target VLM's own prediction errors as \emph{natural}, \emph{model-specific} counterfactuals with minimal pipeline overhead.
\looseness=-1

\textbf{Hard negative mining.}
Hard negative mining has a long history in metric and contrastive representation learning~\citep{schroff2015facenet,robinson2021hard,xu2024isometric}. In vision--language pretraining, hard negatives are constructed by perturbing captions~\citep{zhang2024contrasting}, generating synthetic image--text pairs~\citep{patel2024tripletclip,lai2024multimodalcf,kamath2024hardpositive}, or filtering false negatives in noisy web data~\citep{byun2024mafa,kwak2025qure,wang2025helip}. A parallel line treats model errors as a useful learning signal: LLMs cannot reliably self-correct without external feedback~\citep{huang2024llmscannot}, motivating critique routing~\citep{gou2024critic} and online self-correction policies~\citep{kumar2025score}; visual self-improvement methods construct preference pairs or reflection trajectories from VLM mistakes for fine-tuning~\citep{ding2025sherlock,cheng2025vlmselfimprove}; and a complementary line uses failures to drive curriculum design and data synthesis~\citep{lee2025selfimprovingtransformers,liang2025sws,zeng2025evaltree}. Theoretically, ICL itself behaves like error-driven learning, with surprising demonstrations exerting stronger effects than common ones~\citep{zhou2025iclerror}. All of these approaches use errors to drive parameter updates. We differ in two ways: hardness is defined by the target VLM's own predictions, making the pool model-specific; and hard negatives are injected directly into the in-context prompt rather than used to drive parameter updates, surfacing targeted contradictions at inference time without any fine-tuning or external verifiers.
\section{Model Errors Reveal Reusable Failure Modes}
\label{sec:analysis}

\begin{figure}[!t]
\centering
\vspace{-3mm}
\includegraphics[width=\columnwidth]{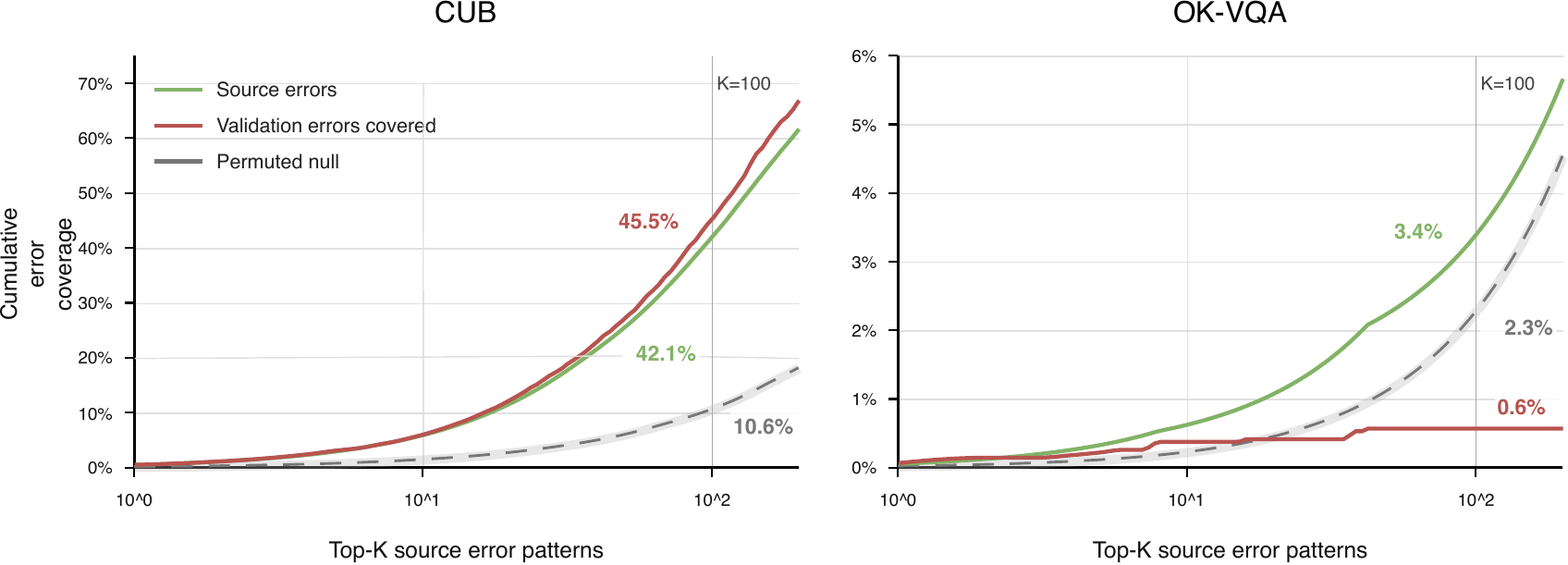}
\caption{\textbf{Exact confusion-pair reuse is concentrated for classification but sparse for VQA.} \textbf{\textcolor{fig_green}{Green}}: source-corpus exact error-pair distribution; \textbf{\textcolor{fig_red}{Red}}: validation errors covered by the top-$K$ most frequent source pairs; \textbf{\textcolor{fig_grey}{Grey}}: permuted-prediction null.}
\label{fig:confusion}
\vspace{-5mm}
\end{figure}

Our central assumption is that a wrong answer can still be informative. When a VLM predicts \emph{fox sparrow} for an image of a \emph{song sparrow}, or answers \emph{circle} when the target answer is \emph{round}, the prediction reveals the answer region the model considered plausible. If similar mistakes recur, then a labeled example the model previously answered incorrectly can serve as a useful warning at test time: it shows the model a case where its likely answer was tempting but wrong.

We test this premise through two questions. First, do VLM mistakes repeat as stable ground-truth/prediction confusions? Second, when exact answer strings do not repeat, do wrong predictions still point toward the right semantic region? These two cases match the two regimes our method must handle: closed-set classification and open-ended VQA.

\begin{hmntakeaway}{Takeaway: Classification mistakes repeat exactly; VQA mistakes repeat semantically}
Classification errors often repeat exactly: the same true class is confused with the same wrong class on unseen examples. VQA errors rarely repeat exactly, but wrong answers still land near the correct answer in semantic space.
\end{hmntakeaway}

\textbf{Closed-set errors repeat as exact confusions.} We first test the strongest form of reuse: whether the same ground-truth answer is repeatedly confused with the same wrong prediction. For each wrong zero-shot prediction, we record an error pair $(y,\hat{y})$, where $y$ is the ground truth and $\hat{y}$ is the model prediction. On fine-grained classification, exact confusions are highly reusable: the top-$100$ source error pairs cover $45.5\%$ of validation mistakes, and $94.2\%$ of wrong validation predictions repeat a pair seen in the source corpus (\Cref{fig:confusion}). This is the cleanest case for hard negatives: the model repeatedly falls into the same decision-boundary traps.

\textbf{Open-ended VQA needs semantic rather than exact reuse.} In VQA, exact strings fragment across synonyms, paraphrases, and partially specified answers. Only $3.0\%$ of wrong validation predictions repeat an exact source pair, so exact pair identity is too brittle. However, this does not mean the errors are arbitrary. The relevant question is whether a wrong prediction still lands near the correct answer semantically. We test this by embedding incorrect predictions and their ground-truth answers, and comparing their cosine similarities against a random ground-truth baseline.

\begin{figure}[!t]
\centering
\includegraphics[width=\textwidth]{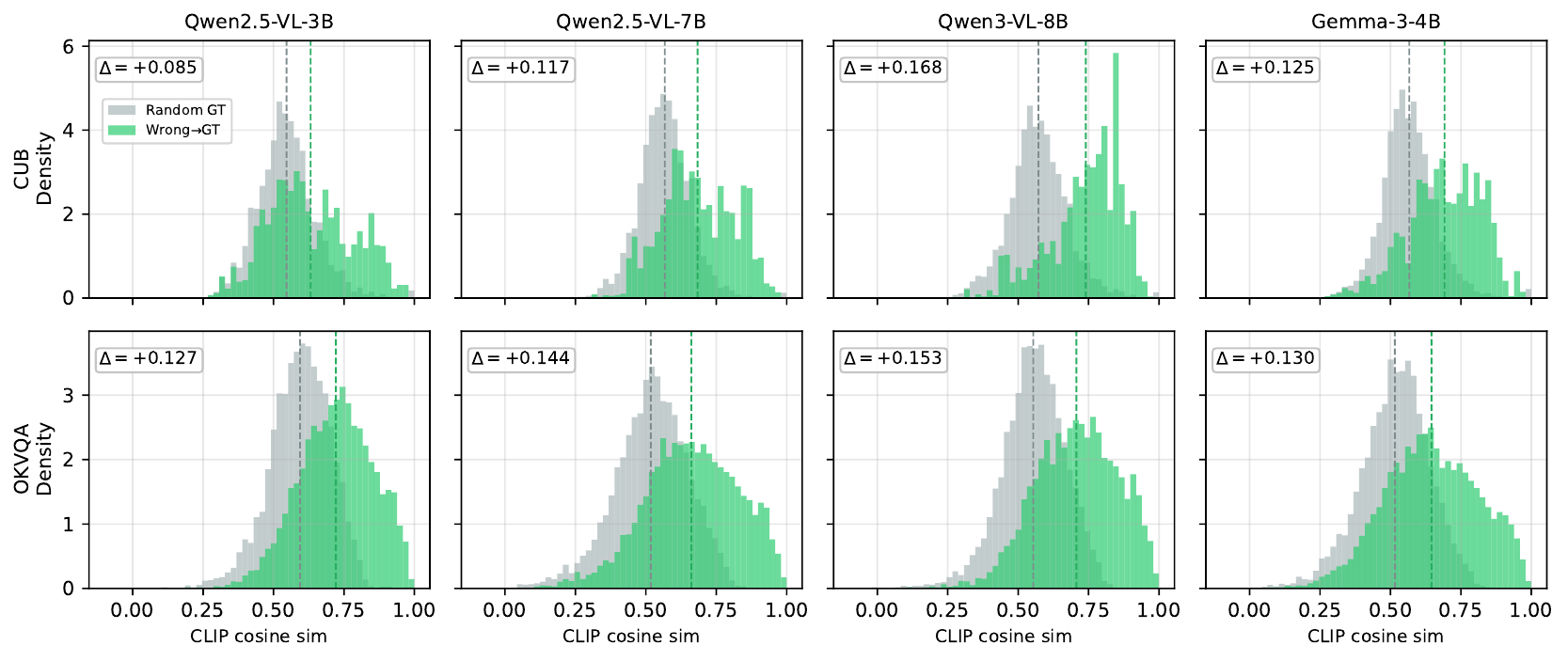}
\caption{\textbf{Wrong predictions remain semantically closer to the ground truth than random answers.} Rows are task regimes; columns span model size/generation/family. Every cell shows the wrong-prediction distribution shifted to the right of the random-GT baseline, with a positive gap $\Delta$.}
\label{fig:semantic-direction}
\end{figure}

\textbf{Wrong predictions tend to concentrate near the truth.} Across all VLMs and both task regimes, wrong predictions are consistently closer to the ground truth than random answers (\Cref{fig:semantic-direction}). For the reference model, the mean gap is $\Delta{=}{+}0.117$ on classification and $\Delta{=}{+}0.144$ on VQA; across all models, every gap is positive. Qualitatively, the mistakes usually preserve the coarse answer type: bird species are confused with visually related species, and VQA answers often differ by synonym, granularity, or category membership. The model is wrong, but it is usually wrong in a meaningful neighborhood. Full counts and significance tests are in \Cref{app:error-structure}.

\textbf{Design implication.} These results show that a mined hard negative is not just a failed example; it is a labeled example attached to a known likely mistake $\hat{y}_h$. At test time, the query's zero-shot prediction $\hat{y}_q$ reveals the model's current likely mistake. Matching $\hat{y}_q$ to stored mistakes therefore gives \method{} a direct way to select examples from the relevant failure mode: exact confusion matching is natural for classification, while semantic matching is necessary for VQA.

\section{Method}
\label{sec:method}

\subsection{Problem Setup}

Let $\mathcal{K}=\{(I_j,Q_j,y_j)\}_{j=1}^M$ denote a labeled corpus of images, questions, and answers, and let $\Phi$ denote a frozen target VLM. Given a test query $(I_q,Q_q)$, our goal is to construct an in-context demonstration set that both supports the correct answer and warns the model away from its most likely error. We use a frozen embedding model $E=(\mathrm{Emb}_I, \mathrm{Emb}_T)$, where $\mathrm{Emb}_I$ embeds images and $\mathrm{Emb}_T$ embeds text. The framework is agnostic to the particular embedding model; the only requirement is that $E$ provide comparable image and text representations for similarity-based selection.

\subsection{Framework Overview}
\label{sec:framework}

\begin{figure}[t]
\centering
\includegraphics[width=\textwidth]{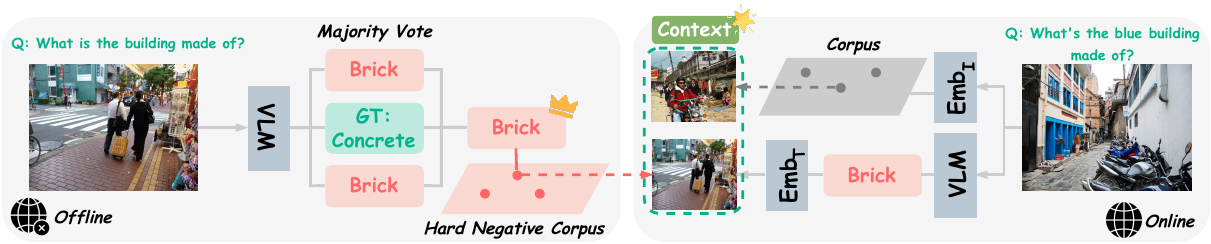}
\caption{\textbf{\method{} mines failures offline and reuses them online as counterfactual demonstrations.} Left: the target VLM is run on the labeled corpus; examples whose voted prediction disagrees with the ground truth are stored with both the gold answer and the mistaken answer. Right: for a new query, image similarity retrieves positive demonstrations, while the query's zero-shot prediction is matched in text-embedding space to stored mistaken predictions; the resulting hard negatives are appended to the context before final inference.}
\label{fig:method}
\end{figure}

\Cref{fig:method} gives an overview. \method{} decomposes the in-context demonstrations into two complementary branches:
\begin{align}
\mathcal{P}_q &= \mathrm{TopK}_{x \in \mathcal{K}}\ \cos\!\big(\mathrm{Emb}_I(I_q),\, \mathrm{Emb}_I(I_x)\big), \label{eq:positive} \\
\mathcal{N}_q &= \mathrm{TopN}_{h \in \mathcal{H}}\ \cos\!\big(\mathrm{Emb}_T(\hat{y}_q),\, \mathrm{Emb}_T(\hat{y}_h)\big), \label{eq:negative}
\end{align}
where $\mathcal{P}_q$ is a set of $K$ positive demonstrations identified by image-embedding similarity, $\mathcal{H}$ is a mined hard-negative pool (\Cref{sec:mining}), $\hat{y}_q$ is the model's zero-shot prediction for the query, and $\hat{y}_h$ is the mistaken prediction stored with each hard negative. The final prompt concatenates $\mathcal{P}_q$ and $\mathcal{N}_q$ before the query, with the hard-negative block labeled as ``tricky examples that were previously confused with the predicted answer''; any hard negative already in the positive set is excluded.

\subsection{Offline Hard-Negative Mining}
\label{sec:mining}

The hard-negative pool $\mathcal{H}$ is constructed once per target model. For every labeled example $(I_j,Q_j,y_j)$, we run zero-shot inference with self-consistency voting~\citep{wang2023selfconsistency,moCICLContrastiveIncontext2024}:
\begin{equation}
\hat{y}_j = \mathrm{vote}\big(\Phi(I_j,Q_j)^{(1)}, \ldots, \Phi(I_j,Q_j)^{(R)}\big).
\label{eq:vote}
\end{equation}
If $\hat{y}_j \neq y_j$, we add the example to the pool:
\begin{equation}
\mathcal{H} = \bigl\{(I_j,Q_j,y_j,\hat{y}_j)\ \big|\ \hat{y}_j \neq y_j\bigr\}.
\label{eq:pool}
\end{equation}

Each mined example records both the ground-truth answer $y_j$ and the model's wrong prediction $\hat{y}_j$. When shown in context, a hard negative's ground-truth answer contradicts the model's likely prediction, which is the core function of a counterfactual demonstration. Mining is model-specific by design: different VLMs have distinct failure profiles, and mining separately ensures the pool captures each model's unique error patterns. We empirically validate this choice and characterize how far a pool mined for one VLM can be reused on another in \Cref{sec:cross-model-main}.

\subsection{Prediction-Conditioned Selection}
\label{sec:selection}

At test time, we first identify $K$ positive demonstrations with image-embedding similarity (\Cref{eq:positive}), following the standard RICES retrieval setup~\citep{yang2022empirical,awadalla2023openflamingo}. We then obtain a zero-shot prediction $\hat{y}_q = \Phi(I_q, Q_q)$ for the query and use it as a \emph{direction indicator} for hard-negative selection.

The selection is performed in text-embedding space. We encode $\hat{y}_q$ and each mined $\hat{y}_h$ with $\mathrm{Emb}_T$ and rank hard negatives by cosine similarity (\Cref{eq:negative}). The top $N$ hard negatives, those whose mistaken predictions are most semantically similar to the query's predicted answer, are selected as contrastive demonstrations.

\textbf{Why semantic text matching?} We use embedding similarity rather than exact matching or token F1 because open-ended VQA predictions exhibit high lexical variation (e.g., ``park,'' ``public garden,'' ``green space'' for the same error mode); see \Cref{tab:ablation-selection}.

\section{Experiments}
\label{sec:experiments}

Our experiments are organized around three research questions. \textbf{RQ1}: Does \method{} improve visual ICL across task regimes, model families, and metrics? \textbf{RQ2}: Are mined mistakes useful because they contain reusable task signals rather than simply adding more examples? \textbf{RQ3}: Which parts of prediction-conditioned hard-negative selection are necessary?

\begin{table*}[!t]
\centering
\caption{\textbf{Main comparison with both primary metric and F1 reported per dataset.} The \colorbox{box_green!12}{\textcolor{box_green}{best}} and \colorbox{box_red!12}{\textcolor{box_red}{second best}} scores per model and metric are highlighted.}
\label{tab:main-comparison}
\resizebox{\textwidth}{!}{%
\begin{tabular}{llrrrrrrrrrr}
\toprule
\multirow{2}{*}{Model} & \multirow{2}{*}{Method}
  & \multicolumn{2}{c}{CUB} & \multicolumn{2}{c}{Flowers}
  & \multicolumn{2}{c}{OK-VQA} & \multicolumn{2}{c}{VizWiz}
  & \multicolumn{2}{c}{Average} \\
\cmidrule(lr){3-4}\cmidrule(lr){5-6}\cmidrule(lr){7-8}\cmidrule(lr){9-10}\cmidrule(lr){11-12}
& & \multicolumn{1}{c}{Acc} & \multicolumn{1}{c}{F1}
& \multicolumn{1}{c}{Acc} & \multicolumn{1}{c}{F1}
& \multicolumn{1}{c}{EM} & \multicolumn{1}{c}{F1}
& \multicolumn{1}{c}{EM} & \multicolumn{1}{c}{F1}
& \multicolumn{1}{c}{Score} & \multicolumn{1}{c}{F1} \\
\midrule\midrule
\multirow{7}{*}{\mbox{Gemma3-4B}}
  & None    & 10.56 & 8.19 & 46.76 & 46.50 & 19.12 & 26.70 & 39.85 & 57.26 & 29.07 & 34.66 \\
  & Random  & 9.98 & 8.10 & 37.01 & 39.51 & 24.81 & 30.74 & 51.84 & 67.08 & 30.91 & 36.36 \\
  & RICES   & 65.40 & 67.62 & 86.70 & 87.43 & 26.65 & 32.72 & 56.08 & 70.40 & 58.71 & 64.54 \\
  & MUIER   & 65.21 & 67.42 & 86.39 & 87.28 & 26.87 & 32.82 & 56.59 & 70.75 & 58.77 & 64.57 \\
  & MMICES  & 13.95 & 10.98 & 38.09 & 38.20 & 26.24 & 32.41 & 52.72 & 68.21 & 32.75 & 37.45 \\
  & \circles & \secondscore{71.97} & \secondscore{72.39} & \secondscore{93.32} & \secondscore{93.49} & \secondscore{31.27} & \secondscore{36.89} & \secondscore{57.61} & \secondscore{71.35} & \secondscore{63.54} & \secondscore{68.53} \\
  & \methodtable & \bestscore{72.22} & \bestscore{72.75} & \bestscore{96.89} & \bestscore{96.88} & \bestscore{35.01} & \bestscore{39.94} & \bestscore{69.69} & \bestscore{76.20} & \bestscore{68.45} & \bestscore{71.44} \\
\midrule
\multirow{7}{*}{\mbox{Gemma3-12B}}
  & None    & 30.34 & 25.02 & 67.51 & 64.33 & 25.47 & 33.89 & 56.82 & 71.26 & 45.04 & 48.62 \\
  & Random  & 29.27 & 25.82 & 71.28 & 71.56 & 33.59 & 39.56 & 70.13 & 80.54 & 51.07 & 54.37 \\
  & RICES   & 76.37 & 76.25 & 96.44 & 96.29 & 36.86 & 42.61 & 73.98 & \secondscore{83.43} & 70.91 & 74.65 \\
  & MUIER   & 76.51 & 76.39 & 96.42 & 96.28 & 36.58 & 42.30 & 73.72 & 83.04 & 70.81 & 74.50 \\
  & MMICES  & 36.37 & 31.61 & 71.36 & 69.13 & 35.12 & 40.85 & 71.50 & 81.56 & 53.59 & 55.79 \\
  & \circles & \bestscore{77.03} & \bestscore{76.90} & \secondscore{97.77} & \secondscore{97.75} & \secondscore{37.75} & \secondscore{43.23} & \secondscore{74.30} & 82.35 & \secondscore{71.71} & \secondscore{75.06} \\
  & \methodtable & \secondscore{76.98} & \secondscore{76.89} & \bestscore{98.85} & \bestscore{98.86} & \bestscore{43.67} & \bestscore{48.29} & \bestscore{78.58} & \bestscore{84.40} & \bestscore{74.52} & \bestscore{77.11} \\
\midrule
\multirow{7}{*}{\mbox{Qwen2.5-VL-3B}}
  & None    & 7.09 & 5.97 & 22.57 & 24.04 & 42.29 & 46.12 & \secondscore{75.83} & \secondscore{78.58} & 36.95 & 38.68 \\
  & Random  & 15.81 & 15.63 & 41.76 & 44.57 & 41.54 & 45.30 & 74.48 & 77.40 & 43.40 & 45.73 \\
  & RICES   & 72.26 & 73.88 & 93.06 & 92.84 & 42.57 & 46.26 & 70.80 & 73.37 & 69.67 & 71.59 \\
  & MUIER   & 72.25 & 73.87 & 93.06 & 92.83 & 41.66 & 45.82 & 71.61 & 74.74 & 69.64 & 71.81 \\
  & MMICES  & 17.26 & 15.38 & 33.84 & 35.47 & 39.32 & 42.95 & 73.23 & 76.21 & 40.91 & 42.50 \\
  & \circles & \secondscore{74.89} & \secondscore{76.34} & \secondscore{94.70} & \secondscore{94.77} & \secondscore{43.24} & \secondscore{47.27} & 72.93 & 75.33 & \secondscore{71.44} & \secondscore{73.43} \\
  & \methodtable & \bestscore{76.75} & \bestscore{77.75} & \bestscore{94.79} & \bestscore{95.25} & \bestscore{43.75} & \bestscore{47.53} & \bestscore{78.26} & \bestscore{82.72} & \bestscore{73.39} & \bestscore{75.81} \\
\midrule
\multirow{7}{*}{\mbox{Qwen2.5-VL-7B}}
  & None    & 14.83 & 11.79 & 43.47 & 42.14 & 33.31 & 38.33 & 68.70 & 73.30 & 40.08 & 41.39 \\
  & Random  & 26.27 & 24.42 & 47.15 & 49.74 & 41.66 & 46.20 & 76.11 & 78.34 & 47.80 & 49.67 \\
  & RICES   & 82.15 & 81.91 & 98.83 & 98.85 & 43.66 & 48.32 & 73.79 & 76.13 & 74.61 & 76.30 \\
  & MUIER   & 82.14 & 81.90 & 98.93 & 98.95 & \secondscore{44.29} & \secondscore{48.76} & 74.67 & 77.08 & 75.01 & 76.67 \\
  & MMICES  & 34.57 & 29.72 & 49.15 & 47.62 & 43.40 & 47.66 & 70.69 & 72.77 & 49.45 & 49.44 \\
  & \circles & \secondscore{82.17} & \secondscore{82.13} & \secondscore{98.99} & \secondscore{99.04} & 43.54 & 48.53 & \secondscore{77.63} & \secondscore{82.36} & \secondscore{75.58} & \secondscore{78.02} \\
  & \methodtable & \bestscore{82.55} & \bestscore{82.30} & \bestscore{99.33} & \bestscore{99.37} & \bestscore{47.80} & \bestscore{51.61} & \bestscore{83.42} & \bestscore{87.49} & \bestscore{78.28} & \bestscore{80.19} \\
\bottomrule
\end{tabular}
}
\end{table*}

\subsection{Experimental Setup}

\textbf{Datasets.} We evaluate on four standard visual ICL benchmarks: \textbf{CUB-200-2011}~\citep{wah2011cub} (fine-grained bird classification), \textbf{Flowers-102}~\citep{nilsback2008flowers} (flower classification), \textbf{OK-VQA}~\citep{marino2019okvqa} (knowledge-based VQA), and \textbf{VizWiz}~\citep{gurari2018vizwiz} (real-world VQA, ${\sim}32\%$ unanswerable). Classification reports accuracy / weighted F1; VQA reports exact match / token F1.

\textbf{Models.} We evaluate four open VLMs from two families: Gemma-3~\citep{gemmateam2025gemma3} (4B, 12B) and Qwen2.5-VL~\citep{bai2025qwen25vl} (3B, 7B). Additional results on Qwen3-VL~\citep{bai2025qwen3vltechnicalreport} (4B, 8B) are reported in \Cref{app:additional-backbones}. Unless otherwise specified, all analyses, ablations, and probes use Qwen2.5-VL-7B as the default backbone.

\textbf{Baselines.} We compare against \textbf{None} (zero-shot), \textbf{Random}, \textbf{RICES}~\citep{awadalla2023openflamingo}, \textbf{MUIER}~\citep{luo2024textual}, \textbf{MMICES}~\citep{chen2023mmices}, and \textbf{\circles}~\citep{xiong2026circles}. Following~\citet{xiong2026circles}, all baselines use a total budget of $32$ in-context examples, and we directly report the results from \circles.

\textbf{Implementation.} Hard-negative pools are mined offline once per target VLM with self-consistency voting. \method{} uses $K{+}N{=}12$ in-context examples for classification ($K{=}8$, $N{=}4$) and $20$ for VQA ($K{=}12$, $N{=}8$). All experiments report means over $3$ runs; full setup, per-model settings, and dataset statistics are in \Cref{app:implementation}.

\subsection{Main Results}
\label{sec:main-results}

\Cref{tab:main-comparison} presents the main comparison of \method{} against six in-context learning baselines across four datasets and four VLM backbones, reporting both the primary metric (accuracy for classification; exact match for VQA) and the corresponding F1 score. From the results, we observe that: 

\textbf{\method{} consistently outperforms all baselines.} It achieves the highest average task score and average F1 for every backbone, winning against the strongest counterfactual baseline \circles{} with an average task-score gain of $+3.09$ points, and improving over the strongest similarity-only baseline (RICES or MUIER) by $+3.3$ to $+9.7$ points. Notably, \method{} achieves this with only $12$--$20$ in-context examples while all baselines use $32$, so the gain is not a function of longer context. 

\textbf{Gains are largest on open-ended VQA.} \method{} delivers its biggest improvements over \circles{} on VQA benchmarks, where plausible wrong answers are diverse; classification gains are smaller because visual neighbors already provide a strong label signal, and accuracy is closer to its ceiling. 

\textbf{The effect is consistent across model scales and families.} \method{} improves over the SOTA method by $+4.91$, $+2.81$, $+1.95$, and $+2.70$ average task-score points on Gemma and Qwen models with different sizes. Additional results on Qwen3-VL are reported in \Cref{app:additional-backbones}.

\begin{figure}[t]
\centering
\includegraphics[width=\textwidth]{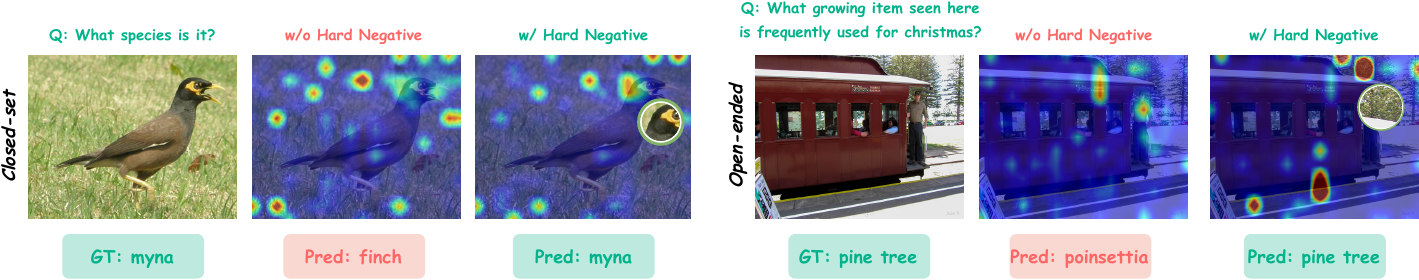}
\caption{\textbf{Hard negatives concentrate attention on the discriminating region and flip the prediction.} Attention from the final prompt token to query-image patches, with vs.\ without the prediction-conditioned hard-negative block.}
\label{fig:attention-heatmap}
\end{figure}

\subsection{Why Mined Errors Help}
\label{sec:pool-decomposition}

The main results show that mined errors improve performance; we now ask why. To probe the mechanism, we (a) partition the CUB labeled corpus by the model's own behavior into a \emph{clean pool} ($1{,}113$ images it classifies correctly under self-consistency voting) and a \emph{hard pool} ($4{,}881$ images it gets wrong), then re-run RICES retrieval restricted to each pool (\Cref{tab:pool-decomposition}); and (b) extract attention from the final prompt token to the query-image patch grid, averaged over heads at late decoder layers, with vs.\ without the hard-negative block (\Cref{fig:attention-heatmap}). From these analyses, we observe that:

\begin{wrapfigure}[13]{r}{0.41\textwidth}
\centering
\captionof{table}{\raggedright \textbf{Pool decomposition} on CUB. \emph{Uncovered}: the query's true class has no clean examples. \emph{Fair}: both pools contain $\geq 4$ examples of the true class.}
\label{tab:pool-decomposition}
\setlength{\tabcolsep}{3.5pt}
\begin{tabular}{l c c c}
\toprule
& Full                & Uncovered           & Fair \\
\midrule\midrule
Full corpus                      & \bestscore{82.15}   & \secondscore{76.27} & \secondscore{89.77} \\
Hard only            & \secondscore{80.34} & \bestscore{80.00}   & 83.27 \\
Clean only          & 46.12               & 11.66               & \bestscore{92.42} \\
\bottomrule
\end{tabular}

\end{wrapfigure}
\textbf{Hard examples are load-bearing precisely on the classes the model gets wrong.} Restricting RICES to the smaller clean pool collapses overall EM from $82.15$ to $46.12$, yet restricting it to the much larger hard pool costs only $1.81$ points. The asymmetry comes from a coverage gap: $77$ of CUB's $200$ species have zero clean examples because the model misclassified every image of those classes, and on the $39\%$ of queries drawn from these classes, hard-only retrieval beats clean-only by $\mathbf{+68.34}$ EM, since clean retrievals are structurally off-class. On the fair slice where both pools cover the true class, clean-only beats hard-only by $\mathbf{+9.15}$ EM, confirming that hard examples are not intrinsically better, but are the only available positives precisely where the model needs them. See \Cref{app:pool-decomposition} for full slice definitions.

\textbf{Visual similarity implicitly performs hard-example coverage.} On average $9.74$ of the top-$12$ RICES retrievals are drawn from the hard pool, because the visually most-confusable images are exactly the ones the model misclassifies; the very property that makes ``hard'' examples seem risky as demonstrations is what makes them naturally retrievable. This motivates \method{}'s two-axis design: positives are retrieved from the full labeled corpus, while the hard-negative branch operates on semantic similarity between predicted answers $\hat{y}_h$ and $\hat{y}_q$, so the two branches provide complementary signals rather than duplicating each other.

\textbf{Hard negatives redirect attention to the discriminating region and flip the prediction.} With the positive set fixed and only the hard-negative block toggled, attention sharpens away from distractor regions and onto the feature that separates the correct answer from the model's plausible mistake. The closed-set and open-ended examples are consistent with the mechanism in \Cref{sec:selection}: a hard negative tied to the model's likely mistake forces it to seek query-image evidence that distinguishes the right answer from that mistake.

\subsection{Cross-Model Hard-Negative Pool Transfer}
\label{sec:cross-model-main}

\begin{table*}[t]
\centering
\caption{\textbf{Cross-model hard-negative pool transfer heatmap.} Each cell reports the change in CUB accuracy or OK-VQA EM relative to the self-mined pool for the same inference backbone and dataset.}
\label{tab:cross-model-heatmap}
\small
\setlength{\tabcolsep}{2.6pt}
\renewcommand{\arraystretch}{1.18}
\resizebox{0.95\textwidth}{!}{%
\begin{tabular}{lccccc|ccccc}
\toprule
\multirow{2}{*}{HN-pool source}
& \multicolumn{5}{c|}{\textbf{CUB}}
& \multicolumn{5}{c}{\textbf{OK-VQA}} \\
\cmidrule(lr){2-6}\cmidrule(lr){7-11}
& Q2.5-3B & Q2.5-7B & Q3-8B & G-4B & G-12B
& Q2.5-3B & Q2.5-7B & Q3-8B & G-4B & G-12B \\
\midrule\midrule
Qwen2.5-VL-3B
 & \cellcolor{black!10}+0.00 & \cellcolor{box_green!24}+0.77 & \cellcolor{box_red!10}-0.23 & \cellcolor{box_green!36}+1.69 & \cellcolor{box_red!26}-1.20
 & \cellcolor{black!10}+0.00 & \cellcolor{box_green!12}+0.24 & \cellcolor{box_red!32}-1.47 & \cellcolor{box_red!38}-1.98 & \cellcolor{box_red!16}-0.71 \\
Qwen2.5-VL-7B
 & \cellcolor{box_green!12}+0.27 & \cellcolor{black!10}+0.00 & \cellcolor{box_red!8}-0.12 & \cellcolor{box_red!8}-0.05 & \cellcolor{box_red!16}-0.65
 & \cellcolor{box_red!12}-0.38 & \cellcolor{black!10}+0.00 & \cellcolor{box_red!26}-1.19 & \cellcolor{box_red!36}-1.89 & \cellcolor{box_red!18}-0.79 \\
Qwen3-VL-8B
 & \cellcolor{box_green!12}+0.26 & \cellcolor{box_green!14}+0.38 & \cellcolor{black!10}+0.00 & \cellcolor{box_green!18}+0.67 & \cellcolor{box_red!14}-0.58
 & \cellcolor{box_red!8}-0.12 & \cellcolor{box_green!12}+0.29 & \cellcolor{black!10}+0.00 & \cellcolor{box_red!14}-0.50 & \cellcolor{box_green!18}+0.52 \\
Gemma-3-4B
 & \cellcolor{box_green!12}+0.30 & \cellcolor{box_green!28}+1.18 & \cellcolor{box_red!8}-0.09 & \cellcolor{black!10}+0.00 & \cellcolor{box_red!8}-0.02
 & \cellcolor{box_red!12}-0.43 & \cellcolor{box_red!12}-0.36 & \cellcolor{box_red!26}-1.20 & \cellcolor{black!10}+0.00 & \cellcolor{box_red!16}-0.67 \\
Gemma-3-12B
 & \cellcolor{box_green!18}+0.53 & \cellcolor{box_green!8}+0.01 & \cellcolor{box_green!14}+0.32 & \cellcolor{box_green!14}+0.31 & \cellcolor{black!10}+0.00
 & \cellcolor{box_red!10}-0.30 & \cellcolor{box_green!8}+0.04 & \cellcolor{box_red!16}-0.71 & \cellcolor{box_red!32}-1.55 & \cellcolor{black!10}+0.00 \\
\bottomrule
\end{tabular}
}
\end{table*}

\method{} mines a model-specific hard-negative pool, but deployment settings may involve multiple VLMs or frequent model upgrades. We therefore ask whether another model can use a pool mined by one model at inference time. \Cref{tab:cross-model-heatmap} reports the change in performance when replacing each inference model's self-mined pool with a pool mined by another backbone, across five backbones (Qwen2.5-VL-3B/7B, Qwen3-VL-8B, Gemma-3-4B/12B).

Transfer works robustly across architecture families. This supports a practical ``mine once, reuse broadly'' interpretation: the stored wrong predictions $\hat{y}_h$ encode semantic neighborhoods of plausible mistakes that are partly shared across model families. The heatmap also reveals the boundary of transfer, so a self-mined pool remains the safest choice when maximum accuracy is required. However, when mining cost or deployment simplicity matters, an existing pool from a nearby or readily available VLM is a viable substitute.

\subsection{Ablations}
\label{sec:ablation}

\begin{wrapfigure}[9]{r}{0.39\textwidth}
\centering
\captionof{table}{\raggedright Ablation on \textbf{component}.}
\label{tab:ablation-components}
\setlength{\tabcolsep}{3pt}
\renewcommand{\arraystretch}{1.05}
\resizebox{\linewidth}{!}{%
\begin{tabular}{@{}ccc|cc@{}}
\toprule
\textbf{Pos.} & \textbf{Rand.\ HN} & \textbf{Pred.\ HN} & \textbf{CUB} & \textbf{OK-VQA} \\
\midrule\midrule
\textcolor{gray}{\xmark} & \textcolor{gray}{\xmark} & \textcolor{gray}{\xmark} & 14.83 & 33.31 \\
\textcolor{box_green}{\cmark} & \textcolor{gray}{\xmark} & \textcolor{gray}{\xmark} & 81.18 & \secondscore{43.16} \\
\textcolor{box_green}{\cmark} & \textcolor{box_green}{\cmark} & \textcolor{gray}{\xmark} & \secondscore{81.23} & 42.86 \\
\midrule
\textcolor{box_green}{\cmark} & \textcolor{gray}{\xmark} & \textcolor{box_green}{\cmark} & \bestscore{82.55} & \bestscore{47.80} \\
\bottomrule
\end{tabular}
}

\end{wrapfigure}
\textbf{Effect of hard negatives.}
\Cref{tab:ablation-components} separates the positive and contrastive branches. Positive demonstrations alone provide the largest single jump, especially on CUB ($14.83{\to}81.18$ Acc), confirming that nearest visual neighbors remain the backbone of visual ICL. Hard negatives then add further gains. On OK-VQA, prediction-conditioned negatives outperform random negatives ($47.80$ vs.\ $42.86$ EM), showing the value of semantically aligned failures in open-ended answer spaces; on CUB, random negatives remain competitive because broader label coverage is enough in this closed-set setting.

\begin{wrapfigure}{r}{0.37\textwidth}
\centering
\captionof{table}{\raggedright Ablation on \textbf{hard selection}.}
\label{tab:ablation-selection}
\setlength{\tabcolsep}{4pt}
\renewcommand{\arraystretch}{1.1}
\resizebox{0.9\linewidth}{!}{%
\begin{tabular}{@{}lccc@{}}
\toprule
& CUB & OK-VQA & Avg. \\
\midrule\midrule
EM & 82.34 & 40.46 & 61.40 \\
F1 & \secondscore{82.39} & \secondscore{46.65} & \secondscore{64.52} \\
Semantic & \bestscore{82.55} & \bestscore{47.80} & \bestscore{65.18} \\
\bottomrule
\end{tabular}
}

\end{wrapfigure}
\textbf{Selection strategy.}
A central design choice in \method{} is how to match the query's zero-shot prediction to the mined pool. \Cref{tab:ablation-selection} compares three candidates (exact string match, token-F1 overlap, and CLIP text similarity) on CUB and OK-VQA. On the closed-set classification task, all three matchers perform comparably (within $0.25$ Acc) because bird species labels have limited lexical variation. On open-ended OK-VQA, exact-match collapses ($40.46$ EM) while semantic matching dominates ($47.80$, $+7.3$ EM): open-ended predictions fragment across synonyms and paraphrases of the same error mode, and only embedding similarity reliably detects them. Text-embedding similarity, therefore, provides a single mechanism that serves both regimes.

\subsection{Hyperparameter Sensitivity}
\label{sec:hparam}

\begin{figure}[t]
\centering
\includegraphics[width=0.85\linewidth]{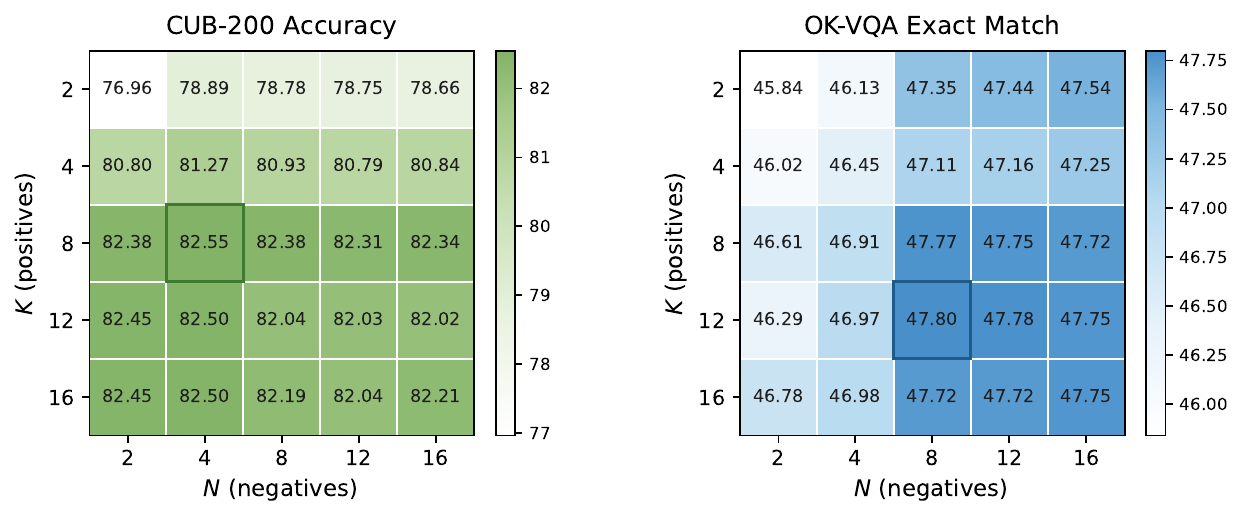}
\caption{\textbf{Hyperparameter sweep} over positive ($K$) and negative ($N$) demonstration counts.}
\label{fig:hparam-heatmap}
\end{figure}

\Cref{fig:hparam-heatmap} sweeps the positive ($K$) and negative ($N$) demonstration counts on CUB and OK-VQA. The two regimes prefer different optima: CUB peaks sharply at $K{=}8$, $N{=}4$, while OK-VQA prefers larger $K$ paired with more negatives ($K{=}12$, $N{\geq}8$) and shows a smooth plateau where any $(K{\geq}8, N{\geq}8)$ combination lies within $0.3$ EM of the best cell. Closed-set classification needs precise visual exemplars with a few targeted contradictions; open-ended VQA benefits from broader context and more diverse contrastive signals. Performance also saturates and eventually degrades when $K{+}N$ exceeds the model's effective context.

\subsection{Inference-Time Cost and Latency}
\label{sec:latency}

\begin{figure}[t]
\centering
\includegraphics[width=0.95\textwidth]{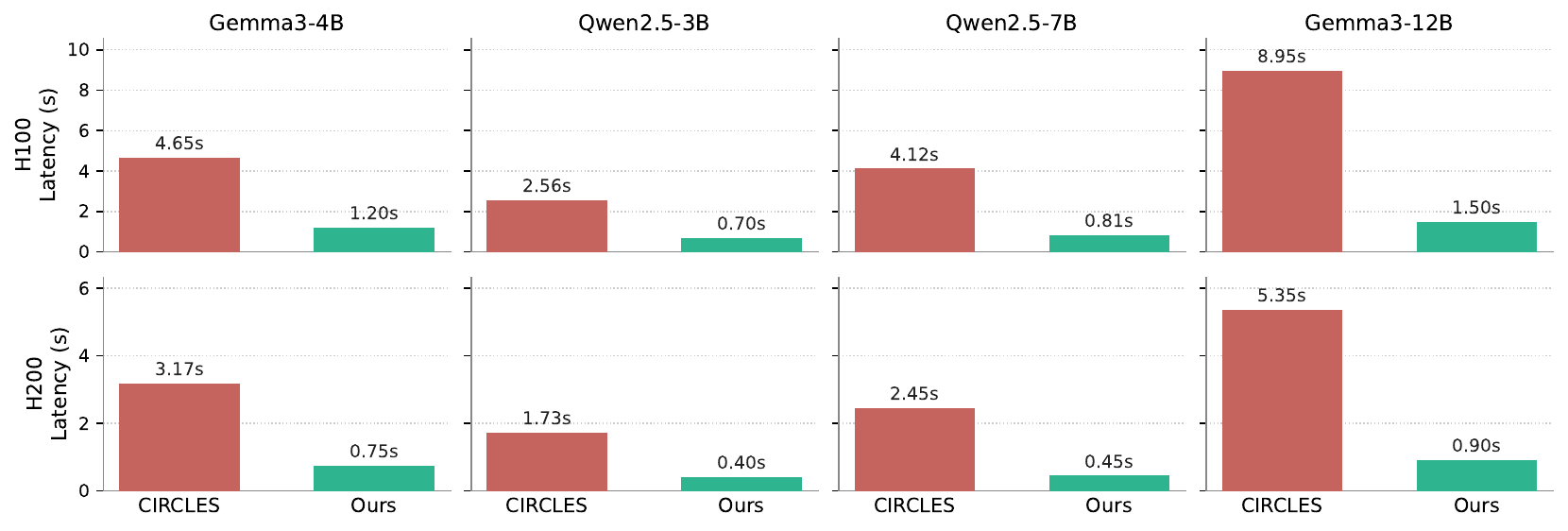}
\caption{\small\textbf{Per-query inference latency} across four VLMs. \method{} is $3.6$--$6.0\times$ faster than \circles{} on H100 (top) and $4.2$--$6.0\times$ faster on H200 (bottom).}
\label{fig:latency-bars}
\vspace{-3mm}
\end{figure}

\method{} separates one-time mining cost from per-query inference cost. Offline mining runs the target VLM over the labeled corpus once; on our largest dataset (VizWiz, ${\sim}20$K examples) this completes in $6.7$--$13.3$ minutes on $8$ H100 GPUs depending on backbone size (full breakdown in \Cref{tab:mining-time} of \Cref{app:mining-time}), and the cross-model transfer above means a previously mined pool can often be reused for a new VLM. At inference time, \method{} adds only a short zero-shot direction call and a nearest-neighbor lookup over pre-encoded predictions ($<\!5\%$ overhead); end-to-end latency is dominated by VLM generation. \Cref{fig:latency-bars} compares both methods inside a shared vLLM pipeline~\citep{xiong2026circles} on H100 and H200: \method{} runs $3.6$--$6.0\times$ faster than \circles{} on H100 and $4.2$--$6.0\times$ on H200, with absolute per-query cost below $1$\,s on every tested backbone on H200 ($0.40$--$0.90$\,s) versus $1.7$--$5.3$\,s for \circles{}. \method{} dominates \circles{} on the accuracy--latency frontier.

\section{Conclusion}
\label{sec:conclusion}

We presented \method, a visual in-context learning framework that replaces attribute-engineered counterfactual selection with model-specific hard-negative mining. By collecting labeled examples that the target VLM incorrectly predicts and selecting a subset whose mistaken predictions are semantically aligned with the query's likely error, \method{} surfaces targeted contradictions that challenge the model's failure modes without requiring attribute extraction, caption generation, or composed image matching. On four benchmarks and four VLMs, \method{} consistently outperforms all compared baselines, including the state-of-the-art counterfactual selection method \circles, with the strongest gains on open-ended VQA where model failure modes are most diverse.

\bibliographystyle{unsrtnat}
\bibliography{references}

\newpage
\appendix
\section*{Appendix}
\startcontents[appendices]
\printcontents[appendices]{l}{1}{\setcounter{tocdepth}{1}}
\vspace{0.4cm}
\section{Implementation Details}
\label{app:implementation}

\textbf{Datasets.}
\textbf{CUB-200-2011}~\citep{wah2011cub} contains $5{,}994$ labeled and $5{,}794$ test images across $200$ fine-grained bird species.
\textbf{Flowers-102}~\citep{nilsback2008flowers} uses $2{,}040$ labeled images and $6{,}149$ test images across $102$ flower categories.
\textbf{OK-VQA}~\citep{marino2019okvqa} contains $9{,}009$ labeled and $5{,}046$ validation questions requiring external knowledge.
\textbf{VizWiz}~\citep{gurari2018vizwiz} contains $20{,}523$ labeled and $4{,}319$ validation questions collected by blind photographers, with ${\sim}32\%$ of questions being unanswerable.

\textbf{Backbone VLMs.}
All baselines and \method{} are instantiated with the same backbone VLMs to ensure fair comparison. We use Gemma-3~\citep{gemmateam2025gemma3} with 4B and 12B parameters, Qwen2.5-VL~\citep{bai2025qwen25vl} with 3B and 7B parameters, and Qwen3-VL~\citep{bai2025qwen3vltechnicalreport} with 4B and 8B parameters. The complete baseline matrix in \Cref{tab:main-comparison} uses the four Gemma-3 and Qwen2.5-VL backbones; additional Qwen3-VL results appear in \Cref{app:additional-backbones}. All models are used in their instruction-tuned variants with greedy decoding (temperature $0$) for final answer generation and served via the vLLM library with BFloat16 precision.

\textbf{Baselines.}
We compare against six in-context learning strategies. \textbf{None}: zero-shot, no demonstrations. \textbf{Random}: uniformly sampled demonstrations. \textbf{RICES}~\citep{awadalla2023openflamingo}: image-similarity selection. \textbf{MUIER}~\citep{luo2024textual}: multimodal selection combining image and text similarity. \textbf{MMICES}~\citep{chen2023mmices}: two-stage multimodal selection. \textbf{\circles}~\citep{xiong2026circles}: attribute-guided counterfactual selection via composed image matching. All baselines use CLIP ViT-G/14 as the shared similarity encoder with a total budget of $32$ in-context examples.

\textbf{Similarity encoder.}
For all similarity-based methods (RICES, MUIER, MMICES, \circles, \method), we use CLIP ViT-G/14 (``laion/CLIP-ViT-g-14-laion2B-s12B-b42K'')~\citep{radford2021learning} as the shared image--text encoder. All visual and textual features are L2-normalized before similarity computation.

\textbf{Hard-negative mining setup.}
For each target VLM, we run zero-shot inference on the labeled corpus with self-consistency voting ($R{=}3$ generations per example, temperature $0.7$, majority vote). Examples where the voted prediction does not match the ground-truth answer are collected into the model-specific hard-negative pool. Pool sizes range from ${\sim}20\%$ of the labeled corpus (for strong 7B+ models on easy tasks) to ${\sim}80\%$ (for small models on difficult tasks).

\textbf{Prompting setup.}
We use a unified prompting template across all methods and tasks. The prompt includes the query image and question, followed by in-context demonstrations, and repeats the input at the end. All VLMs use greedy decoding (temperature $0$) for the final answer generation. For classification tasks (CUB and Flowers), we enforce a closed answer set by appending the candidate class names. For VQA tasks (OK-VQA and VizWiz), the model generates answers freely. VizWiz questions may be unanswerable from the image: each question asks the model to answer ``unanswerable'' in that case, and hard negatives whose stored prediction $\hat{y}_h$ is ``unanswerable'' are excluded from the pool. On VizWiz, each in-context example is labeled with its most frequent annotator answer (ignoring answers marked as not confident) instead of the first annotator's answer, or with ``unanswerable'' if the question is not marked answerable, and the final instruction is ``Answer the question using a single word or phrase.'' After generation, an answerability vote replaces the answer with ``unanswerable'' when at least half of the query's $K{+}N$ in-context examples are labeled ``unanswerable''.

\textbf{\Cref{sec:analysis} analysis setup.}
Unless otherwise specified, the analyses in \Cref{sec:analysis} use Qwen2.5-VL-7B with one representative fine-grained classification dataset (CUB-200) and one representative VQA dataset (OK-VQA). For the confusion-pair analysis, morphological near-duplicates (\emph{jean}/\emph{jeans}, \emph{park meter}/\emph{parking meter}) are treated as correct so that the reported pairs focus on semantic errors. This normalization is only an analytical lens: the mining pipeline of \Cref{sec:mining} uses official VQA equality, and applying the same normalization to mining does not change downstream \method\ performance.

\textbf{Compute resources.}
All experiments were conducted on NVIDIA H100 GPUs. Mining the hard-negative pool is fast: on $8$ H100 GPUs with vLLM and self-consistency $R{=}3$, even VizWiz (the largest dataset, $20$K examples) finishes in $6.7$--$13.3$ minutes per backbone; full per-dataset, per-model timings are in \Cref{tab:mining-time} of \Cref{app:mining-time}. Inference-time overhead is also minimal: one additional zero-shot VLM call per query (for the direction prediction) plus a text-embedding similarity lookup, which adds $<5\%$ wall-clock time compared to RICES.

\section{Offline Mining Time Breakdown}
\label{app:mining-time}

\Cref{tab:mining-time} reports wall-clock mining times measured on $8$ NVIDIA H100 GPUs with vLLM serving (BFloat16, default concurrency) and self-consistency $R{=}3$ generations per example. All entries are end-to-end measurements from this configuration. Mining is one-time per backbone and per dataset, and the cross-model transfer result (\Cref{sec:cross-model-main}) means a previously mined pool can often be reused for a new VLM, further amortizing this cost.

\begin{table}[t]
\centering
\caption{\textbf{Offline mining time on $8$ H100 GPUs.} Wall-clock minutes for one full $R{=}3$-vote zero-shot pass over each dataset's labeled corpus.}
\label{tab:mining-time}
\setlength{\tabcolsep}{6pt}
\renewcommand{\arraystretch}{1.05}
\resizebox{1.0\linewidth}{!}{%
\begin{tabular}{lcccccc}
\toprule
Dataset & \# samples & Gemma-3-12B & Gemma-3-4B & Qwen2.5-VL-7B & Qwen2.5-VL-3B \\
\midrule\midrule
CUB-200    & $5{,}994$  & $3.3$  & $1.9$        & $2.4$  & $1.6$ \\
Flowers-102 & $2{,}040$ & $0.9$          & $0.7$           & $0.8$          & $0.6$ \\
OK-VQA     & $9{,}009$  & $5.1$  & $3.2$        & 3.5  & 2.5 \\
VizWiz     & $20{,}523$ & $13.3$  & $6.7$    & $11.3$ & $6.8$ \\
\bottomrule
\end{tabular}
}
\end{table}

Even on the largest corpus (VizWiz, $20{,}523$ examples), mining completes in well under $15$ minutes for every backbone. Smaller datasets like Flowers-102 are essentially free (under one minute). The implication is that \method{}'s offline cost is not a deployment bottleneck: it is incurred once per backbone-dataset pair and is fully amortized across all subsequent test queries.

\section{Additional Error-Structure Analysis}
\label{app:error-structure}

\textbf{Exact confusion-pair reuse.}
For every wrong zero-shot prediction, we record the pair $(y,\hat{y})$ consisting of the ground-truth answer and the model prediction. The source corpus summarizes past mistakes, and the validation split tests whether those mistakes reappear. As a chance baseline, we randomly re-pair ground-truth answers with wrong predictions, which estimates the repetition expected if errors were unrelated to the true answer.

For CUB, the $100$ most frequent source pairs cover $42.1\%$ of $4{,}766$ source error patterns, which is $4.0\times$ the shuffle baseline ($10.6\%$). These pairs transfer strongly to validation: $94.2\%$ of wrong validation predictions repeat some pair seen in the source corpus, and the top-$100$ source pairs cover $45.5\%$ of validation mistakes. For OK-VQA, exact strings are much more fragmented: the top-$100$ source pairs cover $3.4\%$ of $4{,}407$ source error patterns, compared with a $2.3\%$ shuffle baseline, and only $3.0\%$ of wrong validation predictions repeat an exact source pair.

\textbf{Semantic direction of wrong predictions.}
For each wrong prediction, we encode the predicted answer and the ground-truth answer with CLIP ViT-G/14 and compute cosine similarity. The random-GT baseline re-pairs each wrong prediction with an arbitrary answer from the same task. Across all four VLMs and both task regimes in \Cref{fig:semantic-direction}, the wrong-prediction distribution is shifted right of this random baseline; Welch $t$-statistics range from $32.8$ to $64.6$ ($p{<}10^{-100}$ in every cell). For the reference model, the mean gap is $\Delta{=}{+}0.117$ on classification ($0.684$ vs.\ $0.567$) and $\Delta{=}{+}0.144$ on VQA ($0.662$ vs.\ $0.518$); gaps across all eight cells fall in $[+0.085,+0.168]$.

Mid-similarity examples illustrate what the aggregate statistics capture. Classification mistakes often confuse labels within the same visual family, such as \emph{song sparrow} with \emph{fox sparrow} and \emph{brown pelican} with \emph{white pelican}. VQA mistakes are often synonym- or category-level confusions, such as \emph{children} with \emph{kid}, \emph{round} with \emph{circle}, and \emph{apple} with \emph{mac}. These examples are still incorrect under the task metric, but they preserve enough semantic direction to make prediction-conditioned hard-negative selection plausible.

\section{Effect of the Similarity Encoder}
\label{app:retriever-ablation}

\method{} relies on a frozen similarity encoder $E=(\mathrm{Emb}_I,\mathrm{Emb}_T)$ for two operations: image-similarity positive retrieval and text-similarity hard-negative selection. The main experiments use CLIP ViT-G/14 (LAION-2B). \Cref{tab:retriever-ablation} reports \method{} performance on CUB-200 with Qwen2.5-VL-7B as we swap this encoder across seven open alternatives spanning two families (CLIP and SigLIP\,2) and four parameter scales, holding all other settings fixed.

\begin{table}[t]
\centering
\caption{\textbf{Effect of the similarity encoder on \method{} performance.} CUB-200 with Qwen2.5-VL-7B, swapping the encoder used for both positive retrieval and hard-negative selection while keeping all other settings fixed. CLIP ViT-G/14 is the paper baseline used in all other experiments.}
\label{tab:retriever-ablation}
\setlength{\tabcolsep}{6pt}
\renewcommand{\arraystretch}{1.05}
\begin{tabular}{lcc}
\toprule
Encoder & EM & F1 \\
\midrule\midrule
CLIP ViT-B/32            & 70.85          & 70.50 \\
CLIP ViT-B/16            & 74.63          & 74.53 \\
CLIP ViT-L/14            & 79.79          & 79.72 \\
CLIP ViT-H/14          & \textbf{83.88} & \textbf{83.54} \\
CLIP ViT-G/14          & 82.55          & 82.30 \\
SigLIP\,2 base/16-224             & 76.68          & 76.36 \\
SigLIP\,2 large/16-256            & 78.68          & 78.41 \\
\bottomrule
\end{tabular}
\end{table}

Three observations emerge. \emph{First, performance scales smoothly with encoder capacity within a family.} On CLIP, EM rises from $70.85$ at ViT-B/32 to $83.88$ at ViT-H/14, with strict monotonicity in encoder size; the SigLIP\,2 base$\to$large step shows the same direction ($76.68 \to 78.68$). \emph{Second, the paper baseline is near the top.} CLIP ViT-G/14 (LAION-2B) at $82.55$ EM is within $1.3$ points of the best encoder we tested (CLIP ViT-H/14 LAION-2B, $83.88$) and is the second-best entry overall, confirming that our headline numbers are not propped up by a uniquely strong retriever. \emph{Third, \method{} is robust across the full encoder range:} even the smallest CLIP encoder (ViT-B/32) yields $70.85$ EM, far above zero-shot ($14.83$) and the Random baseline ($26.27$) on the same Qwen2.5-VL-7B backbone (\Cref{tab:main-comparison}). The benefit of mining and contrasting model errors is therefore preserved even when the retriever is significantly downsized, suggesting that the dominant source of \method{}'s gain is the prediction-conditioned hard-negative signal rather than encoder strength.

\section{Additional Backbone Results}
\label{app:additional-backbones}

\Cref{tab:additional-backbones} reports \method{} results on two additional Qwen3-VL backbones. We use the same CLIP encoder and task-specific $K,N$ settings as in the main experiments.

\begin{table}[t]
\centering
\caption{\textbf{Additional \method{} results on Qwen3-VL backbones.} CUB and Flowers report accuracy / weighted F1; OK-VQA and VizWiz report exact match / token-level F1.}
\label{tab:additional-backbones}
\setlength{\tabcolsep}{5pt}
\renewcommand{\arraystretch}{1.05}
\resizebox{0.8\linewidth}{!}{%
\begin{tabular}{lcccccccc}
\toprule
\multirow{2}{*}{Model}
 & \multicolumn{2}{c}{CUB}
 & \multicolumn{2}{c}{Flowers}
 & \multicolumn{2}{c}{OK-VQA}
 & \multicolumn{2}{c}{VizWiz} \\
\cmidrule(lr){2-3}\cmidrule(lr){4-5}\cmidrule(lr){6-7}\cmidrule(lr){8-9}
            & Acc   & F1    & Acc   & F1    & EM    & F1    & EM    & F1    \\
\midrule\midrule
Qwen3-VL-4B & 84.08 & 89.42 & 99.45 & 99.46 & 44.15 & 49.03 & 81.94 & 87.95 \\
Qwen3-VL-8B & 83.87 & 89.41 & 99.61 & 99.62 & 47.50 & 51.70 & 82.39 & 87.51 \\
\bottomrule
\end{tabular}
}
\end{table}

\section{Pool Decomposition: Phase Transition with \texorpdfstring{$k$}{k}}
\label{app:pool-decomposition}

This appendix expands the main-paper pool-decomposition (\Cref{sec:pool-decomposition}, \Cref{tab:pool-decomposition}) with formal slice definitions and a phase-transition analysis. All numbers on CUB with Qwen2.5-VL-7B; primary metric is exact match.

\textbf{Pool partition and slice definitions.}
Hard-negative mining via $R{=}3$-vote zero-shot (\Cref{sec:method}) places a CUB labeled image in the hard pool if the model's majority vote disagrees with the gold label, and in the clean pool otherwise. This yields $1{,}113$ clean and $4{,}881$ hard images out of $5{,}994$ total ($81\%$ hard). The class coverage of the two pools is asymmetric: only $123$ of CUB's $200$ species appear in the clean pool, and the remaining $77$ species have \emph{zero} model-correct labeled examples. We define three test-side slices accordingly:
\begin{itemize}
\item \textbf{Uncovered} ($2{,}255$ queries, $39\%$ of test): the query's true class has $0$ examples in the clean pool, so any clean retrieval is structurally off-class.
\item \textbf{Fair $\geq k$} (at $k{=}4$: $1{,}584$ queries, $27\%$ of test): the query's true class has $\geq k$ examples in \emph{both} pools, so clean and hard retrieval can be compared apples-to-apples.
\item \textbf{Full test} ($5{,}794$ queries): no slicing.
\end{itemize}

\textbf{Phase transition with $k$.}
\Cref{tab:fair-subset} shows how the clean-vs-hard gap sharpens as the fair-subset threshold $k$ grows; each row is restricted to classes with $\geq k$ examples in both pools.
\begin{table}[t]
\centering
\caption{\textbf{Fair-subset analysis: queries restricted to classes with $\geq k$ examples in both pools.} Natural wins at small $k$ (best individual visual matches); clean-only dominates at $k{\geq}4$ (consistent class signal beats individual visual quality); hard-only never wins on a fair subset.}
\label{tab:fair-subset}
\setlength{\tabcolsep}{6pt}
\begin{tabular}{c c ccc c}
\toprule
$k$ & queries & natural & hard-only & clean-only & clean$-$hard \\
\midrule\midrule
2  & 2{,}161 & \textbf{90.19} & 83.80 & 86.49 & $+2.69$ \\
4  & 1{,}584 & 89.77 & 83.27 & \textbf{92.42} & $+9.15$ \\
8  & \phantom{1{,}}844   & 89.22 & 79.74 & \textbf{94.55} & $+14.81$ \\
12 & \phantom{1{,}}385   & 87.79 & 78.44 & \textbf{96.10} & $+17.66$ \\
\bottomrule
\end{tabular}
\end{table}

The hard pool is therefore not intrinsically better than the clean pool: it is essential precisely because it covers the classes the model gets wrong, and CLIP image similarity surfaces it implicitly without any need for explicit pool-routing logic at inference time.

\section{Prompt Template}
\label{app:prompt-template}

We adopt a unified prompting template across all backbones and tasks. The \method{} template, shown below, is structured as follows: the query is presented first; $K$ positive demonstrations identified by image-embedding similarity follow; $N$ prediction-conditioned hard negatives are appended under the explicit label \emph{``tricky examples that were previously confused with the predicted answer''}; and the query is repeated at the end for clarity. Each in-context example is rendered as the original \texttt{Image / Question / Answer} triple. For hard negatives we always show the \emph{ground-truth} answer of the labeled example, not the model's stored mistaken prediction $\hat{y}_h$ (the latter serves only as the selection key during retrieval, \Cref{eq:negative}). All VLMs use greedy decoding (temperature $0$); for classification (CUB and Flowers) we additionally enforce a closed-set answer space by appending the candidate class names, while OK-VQA and VizWiz are evaluated open-ended.

\begin{center}
\begin{promptbox}{Prompt template for VLM inference under \method{}}
Your task is to perform \verb|{{Task Type}}|.\\

\verb|{{Query Image}}|\\
Question: \verb|{{Query Question}}|\\

Here are \verb|{{K}}| in-context examples to help you answer the question:\\

\verb|{{Positive Image 1}}|\\
Question: \verb|{{Positive Question 1}}|\\
Answer: \verb|{{Positive Answer 1}}|\\

\verb|{{Positive Image 2}}|\\
......\\

Below are \verb|{{N}}| tricky examples that were previously confused with the predicted answer:\\

\verb|{{Hard-Negative Image 1}}|\\
Question: \verb|{{Hard-Negative Question 1}}|\\
Answer: \verb|{{Hard-Negative Ground-Truth Answer 1}}|\\

\verb|{{Hard-Negative Image 2}}|\\
......\\

Here is the original question again.\\
\verb|{{Query Image}}|\\
Question: \verb|{{Query Question}}|\\

Please provide your response by directly outputting the answer.
\end{promptbox}
\end{center}

\section{Limitations}
\label{app:limitations}

\method{} has several boundaries that we make explicit here.

\textbf{Dependence on the informativeness of the zero-shot prediction.}
The quality of hard-negative selection depends on the query's zero-shot prediction $\hat{y}_q$ being a meaningful error direction. Some predictions will be semantically unrelated to the ground truth, overly generic, or driven by missing world knowledge rather than a local visual or semantic confusion; in these cases, prediction-conditioned selection cannot identify corrective hard negatives because the model has not produced a usable direction to follow. We view this as a boundary of inference-time demonstration selection rather than a flaw specific to \method{}: when the initial prediction is entirely off-manifold, the underlying limitation is likely in the model's internal knowledge or reasoning capability, which may require parameter updates such as supervised fine-tuning or reinforcement learning rather than better demonstration selection alone.

\textbf{Prompt-structure choices are not adaptive.}
Our current implementation uses a fixed prompt structure with a capped number of hard negatives. More sophisticated prompt organizations or adaptive negative budgets that respond to per-query difficulty may yield further improvements but are outside the scope of this paper.

\textbf{Benchmark scope.}
Our evaluation focuses on four benchmarks with available labeled corpora. Extending the approach to settings where examples must be identified from web-scale corpora, or where no labeled pool is available, remains an open direction.

\end{document}